# Cognitive Architecture for Direction of Attention Founded on Subliminal Memory Searches, Pseudorandom and Nonstop

by J. R. Burger

## *History*

The term architecture implies a model not only for input-output behavior but for physical structure. Physical structure refers to size and weight and also to known electrical and logical properties. Neuroscience aside, the central goal of cognitive architecture is to build a practical thinking machine, that is, an intelligent agent that 'reasons' like a human [1]. Towards this end, it is desired to implement cognition in general and not just the detection of a particular signal from the outside world. Preferably, cognition must occur in a timely self-contained way as it does in humans.

Since architecture is supposed to model brain behavior in logical detail, a cognitive architecture must address information readily available in psychology texts, including short term memory. A valid architecture must have memory that is capable of holding for a brief time encoded signals from the senses, emotional signals, and signals created by recall from long term memory.

In a cognitive architecture that models physical structure, subconscious long term memory must be associative; any information may be recalled instantly once proper cues are found in short term memory. Information is committed to subconscious long term memory by a process termed 'rehearsal' in short term memory. We note that gifted individuals have photographic memory that latches instantly, suggesting that long term memory is unrelated to synaptic growth.

## *Background search capabilities*

Everyone has experienced trying to remember something, but being unable to do so. Unknown to you, a search proceeds within your subconscious long term memory for what you are trying to remember. When you least expect it, possibly at an inconvenient moment, the correct information will pop into short term memory with amazing clarity. This is an indication that the brain has an ability to work in the background without being noticed, adding or subtracting cues from a search until the cues are exactly right, as they must be for recall.

Memory searches clearly serve beyond merely remembering the forgotten. There is a theory that decisions are made not by 'free will' but by a search of past similar situations held in memory. Electroencephalography relating to the timing of finger movements, pursued by Benjamin Libit and others, indicates that choices are often made in the brain before a person realizes it [2, 3]. Surprisingly, the brain seems to retain control. In other words, it is not the other way around, in which a 'person' makes a decision, and then tells the brain what to do.

Along similar lines, it may be noted that the brain appears to search itself continuously in the background not only for forgotten facts and situations, but also for solutions to problems. A problem in this context might be a hard problem with no logical solution. Problems like this often use random trial solutions. In analogy, dreams and daydreams are what we experience as the brain attempts to solve difficult or impossible problems by random search. The qualification, of course, is that many difficult problems have no good solutions.

Randomness has advantages. In order to retrieve forgotten memories, including solutions to hard problems based on past experiences, it is efficient for cues to be selected pseudorandomly. Pseudorandomly chosen initial values are commonly used in numerical analysis to solve difficult optimization problems. Since random starting points are helpful in computer science, they might also be helpful for brain memory searches too.

The importance of search suggests an architecture in which a cue editor works tirelessly to recall randomly related information to a subliminal level. Here it undergoes analysis unconsciously. It is noted here that subliminal recalls may occur at a rate of tens per second as permitted by neural circuitry. A person is aware only of the most 'important' recalls, the ones permitted into short term memory. A cognitive architecture is now synthesized that includes pseudorandom memory search and subliminal analysis.

## *Cognitive architecture including subliminal analysis*

Figure 1 illustrates a system of associative memory whose blocks can be related to specific neural circuits that are easily synthesized, since neurons achieve arbitrary Boolean logic [4].

Short term memory neurons may be explained as having digital outputs like any other logical neuron, but their dendritic pulses are a little longer (hundreds of milliseconds) compared to the typical few milliseconds. Short term memory neurons support longer lasting dendritic pulses because of a shortfall of internal potassium or equivalent in their dendrites, causing an extended pulse burst within the axon.

Long term memory neurons may be explained as digital neurons that transfer neurotransmitters from boutons back into dendritic receptors. The result is a digital read-only memory (ROM) neuron that can be set instantly, and will latch indefinitely, as long as the neuron exists, unless cleared because of lack of use. The indefinite cycling of neural signals in a neuron is physically possible because neurons can be modeled as adiabatic, that is, requiring practically no calories for electrical signaling beyond what is required to sustain any biological cell.

Words in memory are assumed wide enough to accommodate every possible feature a human can experience, shape, shade, tone, smell, feel, emotional strength and so on for thousands of elemental features. Each feature is defined by its logical location in a word. Features are encoded from the senses by a large complex neural network labeled the *Sensory Encoder*. If a novel combination of features is used repeatedly, the *Need to*

*Learn* detector calculates that, if possible, the encoder should 'learn' new features. The *need to learn* box may be based on digital filters, as proposed below for memorization enable, although they are not designed here. As an example of learning, the color chartreuse might be learned as a digital AND combination of yellow and green. Combinational learning like this (no analog parameters) is akin to implicit or reflexive memory via new synapses and serotonin; it is assumed driven by a need for efficiency in successful species who cannot afford to waste time first recalling yellow, then green, then a mental definition of chartreuse.

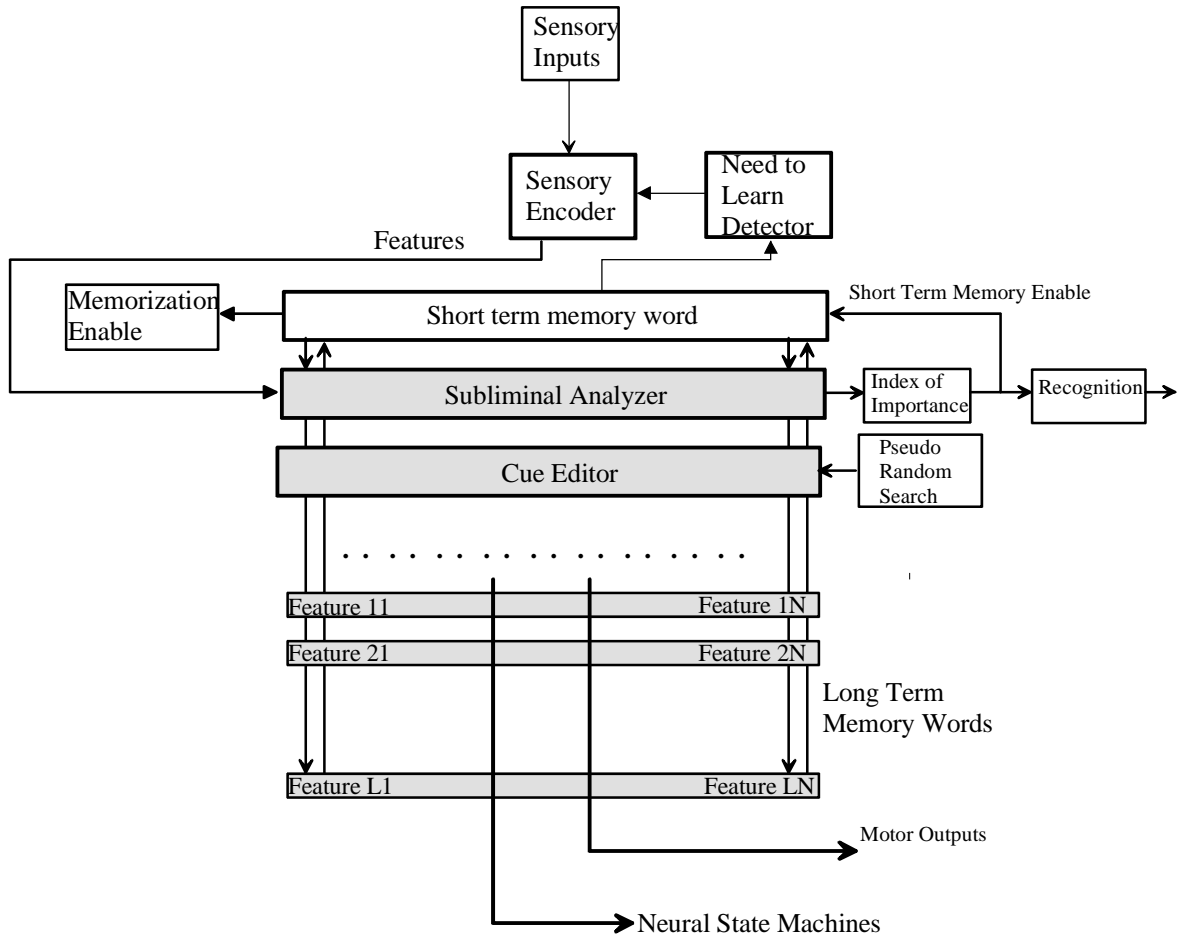

**Figure 1.   Cognitive unconscious architecture including subliminal analysis.**

Certain motor outputs are initiated under feedback control within the brain, giving a sensory response, such as hot coffee on the lips. But there is a type of procedure known as an unconscious procedure that executes automatically. Humans easily execute many such procedures, similar to long poems learned by 'heart' to be performed mindlessly. Once initiated, one simply brushes teeth, drives truck, or adds checkbook without a lot of pondering, as an aid to survival of the human species.

Unconscious procedures are a major aspect of learning, although, as with combinational learning, we cannot yet synthesize neural circuits that enable such learning in practice. Unconscious procedures are conjectured to be the result of interneurons that synapse between words of long term memory, forming a neural state machine. Neural state machines are efficient in that procedural steps avoid passing through the processing associated with short term memory.

## *Long Term memory*

Undeniably, the most important part of learning is subconscious long term memory. Long term memory is more influential than man-made memory with its limited address fields and keywords, in that <u>any</u> subset of features (except a completely empty subset) in short term memory can used as cues to retrieve old images. Those neural paths pointing down in the figure can deliver cues but also serve for applying features to be memorized. Those buses pointing up in the figure bring forth previously memorized features for evaluation in the subliminal analyzer, but only if the cues match exactly.

Multiple matches are dealt with as in any engineering design via a simple neural circuit, so that two separate images cannot be recalled at exactly the same instant. The first recall to reach the subliminal analyzer is assumed to be the one that is evaluated first; other matches are ignored. If the first recall is not correct, meaning its index of importance is relatively low, another subset of cues, slightly different from others in the sequence, is immediately placed on the cue bus as explained below.

The subliminal analyzer is assumed to alternate images from the senses with subconscious recalls. Although the circuitry for this is not shown, it is simple enough. The sensory encoder is disabled while information is being recalled from long term memory. Sensory information in this model is analyzed same as recalls.

## *Cue Editor*

The cue editor in this architecture is envisioned as in Figure 2. All cues are assumed called into and taken from short term memory. But these cues are inconsistent when an image cannot be remembered immediately. So cues are appropriately masked in a pseudorandom way as shown, using a neural shift register counter, typically fed by neural exclusive OR gates. Counters like this can produce a unique subset of cues. Resulting associative recalls will have some correct features, but not necessarily all the right features; recalls can be analyzed many tens per second.

## *Subliminal analyzer*

The analyzer has the task of determining an index of importance for each subliminal set of features. Digital signals from long term memory or the senses appear on interneurons, and are re-encoded as suggested in Figure 3. Note that encoders are not necessarily simple and have yet to be synthesized in a realistic way. Using identical neural circuitry, the digital contents of short term memory are re-encoded into an index of importance; we note that as short term memory fades, importance drops, so new thoughts are expected. At any given time, these encoders assign a digital value to recall-related neural signals. A subliminal image whose index approaches that of current short term memory will be permitted to replace the current contents of short term memory (assumed to be the same as consciousness). In this way, short term memory accomplishes what the theorist Richard Semon termed *direction of attention*.

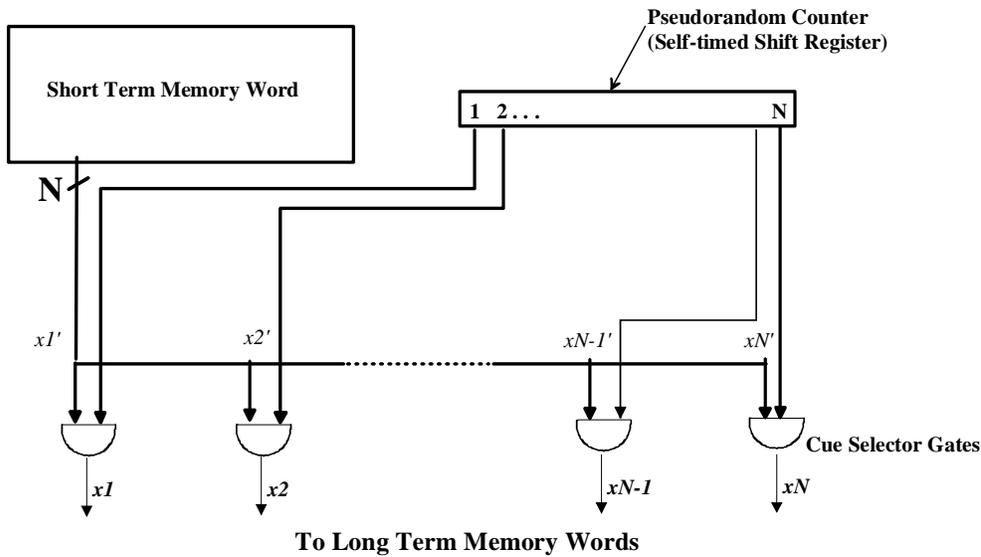

**Figure 2. Cue editor showing pseudorandom counter block.**

Importance is not conflict resolution; it is evaluation for: 1) Brightness of sensory images, 2) Magnitude of emotional content, 3 Quantity of matched cues, and 4) Recency of experiences.

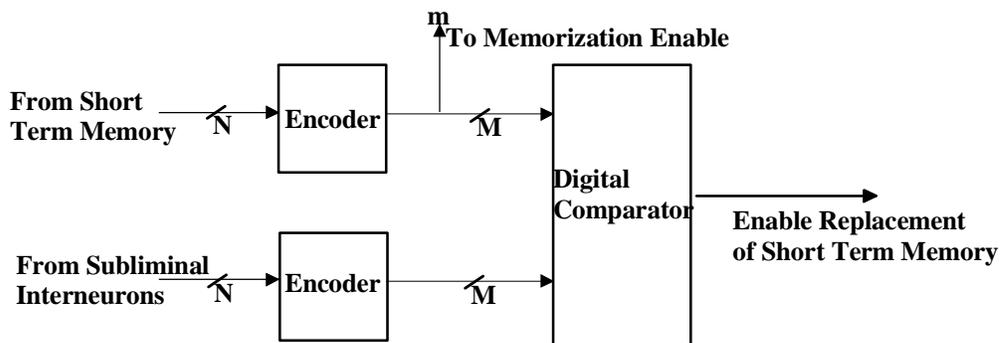

**Figure 3. Calculation of importance.**

Trials proceed at many tens per second. Occasionally there is *recognition*, which is a feeling associated with a particularly clear memory with many matched cues.

## *Memorization enable*

The availability of blank memory words to hold new information is assumed unlimited. Memorization in this architecture is triggered by a *memorization enable* block which is sensitive to recurring images in short term memory, that is, rehearsal.

In the example circuit in Figure 4, conditions for committing something to memory are true if cues are presented but there are no matches, or recalls. Additionally, if a given image, as identified by the above importance encoder, appears in short term memory twice, separated by a given delay, it will be committed to long term memory. The delay can be implemented by short term neurons in a standard digital filter arrangement. A simple neural multi write circuit ensures that only one word is programmed for a given memorization enable.

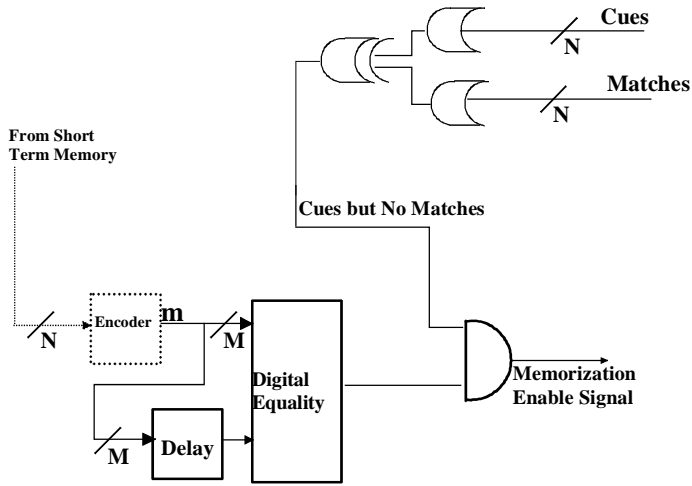

**Figure 4. Proposed memorization enable circuit using digital filter.**

## *Conclusions*

The above cognitive architecture is noteworthy because it includes pseudorandom memory searching as an ongoing process. Cue subsets are selected pseudorandomly, as many as possible per second, so recalls alternate with sensory data, if any, for subliminal analysis. An index of importance is computed for each image by encoders especially for this purpose. When the index of importance for a subliminal image approaches that of current short term memory, a transfer occurs. A new set of attributes is thus enabled to enter short term memory, with a new set of cues, thus defining a new 'thought' and a direction of attention.

Neurons are easily specialized to have short as well as long term memory properties, so, since neurons are capable of arbitrary Boolean logic, and since there are trillions of them, sophisticated digital circuits are possible in the brain. An advantage of the above architecture, from the view of designing an intelligent robot, is that it depends not on meaningless symbols and undefined analog parameters, but rather on neural gates that have hardware equivalents.